\documentclass[10pt, a4paper]{article}
\usepackage[T1]{fontenc}
\usepackage{tipa} % for \textgamma
\usepackage{multirow} 

\DeclareUnicodeCharacter{1E24}{\d{H}}      % uppercase
\DeclareUnicodeCharacter{1E25}{\d{h}}      % lowercase
\DeclareUnicodeCharacter{0263}{\textgamma} % no uppercase
\usepackage[final]{lrec2026} % this is the new style
\title{DATASHI: A Parallel English–Tashlhiyt Corpus for Orthography Normalization and Low-Resource Language Processing.}

\name{Nasser-Eddine Monir$^{1}$, Zakaria Baou$^{2}$}

\address{$^{1}$Université de Lorraine, CNRS, Inria, LORIA, F-54000 Nancy, France\\
        $^{2}$ISIMA, Université Clermont Auvergne, 63178 Aubière, France\\
         nasser-eddine.monir@inria.fr, zakaria.baou@edu.uca.fr\\}

\abstract{
\textbf{DATASHI} is a new parallel English–Tashlhiyt corpus that fills a critical gap in computational resources for Amazigh languages. It contains 5,000 sentence pairs, including a 1,500-sentence subset with expert-standardized and non-standard user-generated versions, enabling systematic study of orthographic diversity and normalization. This dual design supports text-based NLP tasks—such as tokenization, translation, and normalization—and also serves as a foundation for read-speech data collection and multimodal alignment. Comprehensive evaluations with state-of-the-art Large Language Models (GPT-5, Claude-Sonnet-4.5, Gemini-2.5-Pro, Mistral, Qwen3-Max) show clear improvements from zero-shot to few-shot prompting, with Gemini-2.5-Pro achieving the lowest word and character-level error rates and exhibiting robust cross-lingual generalization. A fine-grained analysis of edit operations—deletions, substitutions, and insertions—across phonological classes (geminates, emphatics, uvulars, and pharyngeals) further highlights model-specific sensitivities to marked Tashlhiyt features and provides new diagnostic insights for low-resource Amazigh orthography normalization.
 \\ \newline .}

\begin{document}

\maketitleabstract

\section{Introduction}
\label{sec:introduction}

Tashlhiyt—also known in the literature as Tashelhiyt or Shilha—is one of the three major Amazigh varieties spoken in Morocco, alongside Central Atlas Tamazight and Tarifit. It belongs to the Afroasiatic language family and is primarily used in the southern regions of Morocco, from the High Atlas to the Sous region. Although spoken by several million people, Tashlhiyt remains markedly under-represented in digital resources and computational research. According to recent surveys of Amazigh language technologies \cite{AkallouchEtAl2025}, the overall distribution of natural language processing (NLP) datasets across dialects is highly uneven as Tashlhiyt receives only limited coverage, resulting in a 20–25\% performance gap in multilingual or cross-dialectal systems.

The creation of dedicated computational resources for Tashlhiyt is essential for advancing text-based technologies. Most existing datasets are either too small, thematically narrow, or insufficiently standardized to support robust evaluation and reproducibility. The \textbf{DATASHI} corpus addresses this gap by providing parallel English–Tashlhiyt data capturing both non-standard user-generated writing and expert-standardized equivalents. This dual representation is valuable for developing normalization models and evaluating how orthographic variation affects downstream NLP tasks such as translation, tokenization, and language modeling.

At a broader level, corpora that explicitly encode orthographic diversity not only improve text-based processing, but also facilitate cross-domain transfer to speech-related technologies. Consistent text normalization enhances pronunciation modeling, forced alignment, and lexicon construction—key components for building speech recognition or synthesis systems in low-resource settings. In this sense, while \textbf{DATASHI} is primarily a textual resource, its standardized design enables systematic extension toward read-speech data collection.

This dual purpose aligns with broader efforts in low-resource language documentation, where unified multimodal datasets have proven crucial for accelerating the development of both natural language processing (NLP) and speech technologies~\cite{AddaEtAl2022, Besacier2014}. Recent multilingual initiatives such as \textit{GlobalPhone}~\cite{Schultz2002}, \textit{CMU Wilderness}~\cite{Black2019}, and \textit{BABEL}~\cite{Gales2014} have shown that building balanced, linguistically informed resources across modalities can significantly reduce data scarcity bottlenecks. However, for Amazigh languages—and especially for Tashlhiyt—no comparable resource currently exists that provides clean, normalized text suitable for phoneme-level modeling or cross-modal alignment.

The paper is organized as follows. Section~\ref{sec:background} reviews prior work and key challenges in Tashlhiyt NLP. Section~\ref{sec:corpus} introduces the \textbf{DATASHI} corpus and its linguistic design. Section~\ref{sec:methodology} details the orthographic normalization framework. Section~\ref{sec:experimental} outlines the evaluation setup, and Section~\ref{sec:results} presents results and phonological analyses before concluding with future directions in Section~\ref{sec:conclusion}.

\section{Background}
\label{sec:background}

\subsection{Amazigh and Tashlhiyt NLP}

Moroccan Amazigh (Berber) languages remain among the most under-resourced languages in natural language processing. Existing digital resources are scarce, unevenly distributed, and lack large annotated corpora or standardized processing tools~\cite{FadouaBoulaknadel2012,AkallouchEtAl2025}. Available datasets mainly target part-of-speech tagging, named-entity recognition, and small-scale parallel corpora, but remain insufficient for large-model training and evaluation~\cite{Maarouf2025,AmriZenkouar2017}. 

Computational work has produced a limited number of morphological analyzers, OCR systems for Tifinagh\footnote{Tifinagh is an alphabet of ancient Amazigh origin used to write modern Amazigh languages, officially adopted in Morocco for standardized orthography.}, and experimental machine-translation models, although existing developments remain small-scale, domain-specific, and rarely integrated into end-to-end NLP pipelines~\cite{AkallouchEtAl2025}. Comprehensive surveys consistently identify morphological richness, dialectal heterogeneity, and orthographic variation as primary barriers to progress. Moreover, resource distribution remains asymmetric: Standard Moroccan Tamazight benefits from comparatively larger datasets, while Tashlhiyt and Tarifit varieties remain critically under-represented~\cite{AmriZenkouar2017}.

For Tashlhiyt, linguistic investigations document its particularly complex morphology, including non-concatenative root-and-pattern processes, extensive use of consonant gemination, and intricate agreement systems for gender and number~\cite{Ridouane2014,Oussou2021,Riad2022}. 

Substantial script and orthographic variation within Tashlhiyt—from Latin, Neo-Tifinagh, and Arabic scripts to non-standard user-generated orthographies—renders tasks such as tokenization, sentence alignment, and normalization significantly more difficult than in more standardized languages~\cite{FadouaBoulaknadel2012, ChakerSelles2016}. 

Moreover, available corpora that cover Tashlhiyt (or include it) remain very limited, and high-quality, large-scale parallel corpora aligning Tashlhiyt with a major language are notably absent~\cite{AkallouchEtAl2025}. The development of a dedicated parallel and standardized corpus therefore constitutes an essential step toward enabling reproducible evaluation, advancing normalization research, and supporting the broader creation of NLP resources for Tashlhiyt and related Amazigh varieties. 

\subsection{Normalization}

\begin{table*}[t]
\centering
\begin{tabular}{lccccc}
\hline
\textbf{Language} & \textbf{Tokens} & \textbf{Avg. sentence len.} & \textbf{Min len.} & \textbf{Max len.} & \textbf{Vocabulary} \\
\hline
English & 43,654 & 8.68 & 2 & 16 & 3,160 \\
Non-std. Tashlhiyt & 35,714 & 7.11 & 2 & 18 & 7,261 \\
\hline
\end{tabular}
\caption{Statistics of the DATASHI Corpus}
\label{tab:corpus-stats}
\end{table*}

\begin{table*}[t]
\centering
\begin{tabular}{lccccc}
\hline
\textbf{Language} & \textbf{Tokens} & \textbf{Avg. sentence len.} & \textbf{Min len.} & \textbf{Max len.} & \textbf{Vocabulary} \\
\hline
Standard Tashlhiyt & 12,211 & 8.14 & 2 & 21 & 3,683 \\
Non-std. Tashlhiyt & 10,698 & 7.13 & 2 & 18 & 4,583 \\
\hline
\end{tabular}
\caption{Statistics for the 1,500-sentence Tashlhiyt subset of the DATASHI corpus, comparing non-standard and expert-standardized versions used for orthography normalization.}
\label{tab:subset-stats}
\end{table*}

The orthographic diversity observed in Tashlhiyt reflects the broader multiscriptal situation of Amazigh writing in Morocco. Since the official adoption of the Neo-Tifinagh alphabet by the Royal Institute of Amazigh Culture (IRCAM) in 2003, it has served as the institutional standard for Amazigh literacy. However, Latin-based transcription remains predominant in research, education, and digital communication~\cite{FadouaBoulaknadel2012, ChakerSelles2016}. 

In informal contexts—particularly on social media—users employ both the standard Arabic script and, frequently, Latin orthographies adapted to online conventions. This latter system involves employing numerals and non-standard characters (e.g., “3”, “7”, “9”) to represent phonemes absent from the Latin script.~\cite{Boulaknadel2018}. This hybrid digital orthography, combined with the coexistence of these two scripts with Tifinagh, amplifies inconsistency in written data and hinders the design of unified NLP pipelines. As noted by~\cite{AkallouchEtAl2025}, such script heterogeneity poses structural challenges for text normalization, tokenization, and corpus alignment in low-resource Amazigh settings.

Text normalization—the process of mapping non-standard spellings and orthographic variants to a consistent form—is a critical component of NLP pipelines for low-resource languages. In this regard, normalization has been widely explored as a strategy to mitigate orthographic noise and lexical sparsity. Studies on Arabic dialects, African languages, and South Asian languages demonstrate that explicit normalization significantly improves downstream performance in machine translation, ASR post-processing, and morphological tagging \cite{ZhangEtAl2023,AlSallab2018,Haque2021}.

In the case of Tashlhiyt, the literature still reports very limited work addressing orthographic normalization, either at the word or sentence level~\cite{AkallouchEtAl2025}, leaving a clear methodological and empirical gap that this study seeks to address.

Recent research has investigated large language models (LLMs) for spelling correction and orthographic normalization across diverse languages. Models such as GPT-3, mT5, and BLOOM demonstrate promising zero- and few-shot generalization capabilities for normalization, even in languages absent from their training data \cite{Madaan2023,Anastasopoulos2023}. Prompt-based or instruction-tuned variants have been applied to dialect standardization, grapheme-to-phoneme normalization, and text cleanup in under-represented languages, often outperforming smaller supervised systems in low-data regimes \cite{Madaan2023}. These findings suggest that LLMs encode cross-lingual orthographic regularities that could be leveraged for Tashlhiyt normalization, providing both a methodological baseline and a diagnostic probe for language coverage in multilingual foundation models.

\section{Corpus Creation}
\label{sec:corpus}

\subsection{English and non-standard Tashlhiyt}

\textbf{DATASHI} corpus~\footnote{All resources used in this study—including the full dataset (5,000 sentence pairs and the 1,500 standardized subset), preprocessing and evaluation code, and the full prompt text—are publicly available at \url{https://github.com/Nasseredd/datashi} to support transparency and reproducibility.} was designed to capture everyday linguistic diversity in Tashlhiyt while maintaining thematic balance and parallel alignment with English. To achieve this, we defined 20 thematic domains—including \textit{home and daily life, food, health and body, greetings, money, technology,} and \textit{religion}—and generated 250 sentences per domain, yielding a total of 5,000 English sentences. The English sentences were created using a semi-controlled elicitation procedure: initial templates were drafted by the authors and expanded to cover common communicative situations within each domain, then manually reviewed to ensure grammatical correctness, lexical diversity, and naturalness. Each domain was reviewed to ensure representativeness of everyday communicative contexts and coverage of a wide range of lexical and morphosyntactic constructions.

For the parallel alignment, 34 native Tashlhiyt speakers (17 male, 17 female), aged 20–50 years, were recruited. These participants were non-experts, unfamiliar with Latin-based or Neo-Tifinagh orthographies, and were instructed to transcribe the 5,000 English sentences using their own spontaneous non-standard writing conventions. The resulting material therefore reflects authentic variation in community writing practices.

Subsequently, a group of seven Amazigh language experts (four male, three female), all with advanced English proficiency (approximately CEFR level C1 or higher), manually standardized a subset of 1,500 sentences without any prior automatic or semi-automatic normalization step. This subset constitutes the gold-standard reference for the normalization experiments presented later in this study.

Table~\ref{tab:corpus-stats} summarizes corpus-level statistics for the full DATASHI dataset. The English portion comprises 43,654 tokens with an average sentence length of 8.68 tokens (ranging from 2 to 16), while the non-standard Tashlhiyt side contains 35,714 tokens, averaging 7.11 tokens per sentence (range 2–18). The vocabulary size reaches 3,160 unique types in English and 7,261 in non-standard Tashlhiyt, illustrating the high lexical variability induced by unregulated spelling and spacing conventions typical of informal writing.

\subsection{Standard \& non-standard Tashlhiyt}

In addition to the full corpus, Table~\ref{tab:subset-stats} reports statistics for the 1,500-sentence subset used for expert-based standardization. In this subset, the non-standard Tashlhiyt portion contains 10,698 tokens with an average sentence length of 7.13 tokens, while the standardized version includes 12,211 tokens and a slightly higher average length of 8.14 tokens. The maximum sentence length also increases from 18 to 21 tokens after normalization. 

The difference in vocabulary size within the 1,500-sentence subset—between the standard and non-standard versions—is particularly informative. In this subset, the higher lexical diversity observed in the non-standard Tashlhiyt data arises primarily from inconsistent orthographic segmentation. In community writing, speakers often merge or separate morphemes and clitics unpredictably. For example, the expression \texttt{ɣ akud ann} (“At that moment”) can appear as \texttt{ghakudan}, \texttt{ghakudane}, \texttt{gha kudan}, or even \texttt{rakudan}. Such variation inflates the apparent vocabulary size by generating multiple orthographic forms for the same lexical unit. The expert-standardized version mitigates this redundancy, reducing the vocabulary size from 4,583 to 3,683 unique types. 

\begin{table}[t]
\centering

\begin{minipage}{\columnwidth}
\centering
\begin{tabular}{l c}
\hline
\textbf{Phoneme type} & \textbf{Count} \\
\hline
Vowels & 16,101 \\
Consonants & 32,688 \\
\hline
\end{tabular}
\caption{Number of occurrences of vowels and consonants in the standardized Tashlhiyt corpus.}
\label{tab:vowels-consonants}
\end{minipage}

\vspace{0.5em}

\begin{minipage}{\columnwidth}
\centering
\begin{tabular}{l c}
\hline
\textbf{Consonants} & \textbf{Count} \\
\hline
ɣ & 1,193 \\
\textipa{E} & 602 \\
ḥ & 470 \\
x & 327 \\
\hline
\end{tabular}
\caption{Number of occurrences of pharyngeal and uvular consonants in the standardized Tashlhiyt corpus.}
\label{tab:pharyngeals}
\end{minipage}

\vspace{0.5em}

\begin{minipage}{\columnwidth}
\centering
\begin{tabular}{l c}
\hline
\textbf{Consonants} & \textbf{Count} \\
\hline
g\textsuperscript{w} & 82 \\
k\textsuperscript{w} & 42 \\
ɣ\textsuperscript{w} & 13 \\
x\textsuperscript{w} & 6 \\
q\textsuperscript{w} & 4 \\
s\textsuperscript{w} & 2 \\
r\textsuperscript{w} & 1 \\
\hline
\end{tabular}
\caption{Number of occurrences of labialized consonants (\textsuperscript{w}-marked) in the standardized Tashlhiyt corpus.}
\label{tab:labialization}
\end{minipage}

\end{table}

% \begin{table}[t]
% \centering
% \begin{tabular}{l c}
% \hline
% \textbf{Phoneme type} & \textbf{Count} \\
% \hline
% Vowels & 16,101 \\
% Consonants & 32,688 \\
% \hline
% \end{tabular}
% \caption{Number of occurrences of vowels and consonants in the standardized Tashlhiyt corpus.}
% \label{tab:vowels-consonants}
% \end{table}

% \begin{table}[t]
% \centering
% \begin{tabular}{l c}
% \hline
% \textbf{Consonants} & \textbf{Count} \\
% \hline
% ɣ & 1,193 \\
% \textipa{E} & 602 \\
% ḥ & 470 \\
% x & 327 \\
% \hline
% \end{tabular}
% \caption{Number of occurrences of pharyngeal and uvular consonants in the standardized Tashlhiyt corpus.}
% \label{tab:pharyngeals}
% \end{table}

% \begin{table}[t]
% \centering
% \begin{tabular}{l c}
% \hline
% \textbf{Consonants} & \textbf{Count} \\
% \hline
% g\textsuperscript{w} & 82 \\
% k\textsuperscript{w} & 42 \\
% ɣ\textsuperscript{w} & 13 \\
% x\textsuperscript{w} & 6 \\
% q\textsuperscript{w} & 4 \\
% s\textsuperscript{w} & 2 \\
% r\textsuperscript{w} & 1 \\
% \hline
% \end{tabular}
% \caption{Number of occurrences of labialized consonants (\textsuperscript{w}-marked) in the standardized Tashlhiyt corpus.}
% \label{tab:labialization}
% \end{table}

A similar phenomenon is observed in the complete 5,000-sentence corpus, where the non-standard Tashlhiyt side reaches 7,261 unique types. The broader thematic coverage and greater number of contributors amplify orthographic diversity.

The phonemic composition of the standardized Tashlhiyt corpus is summarized in Tables~\ref{tab:vowels-consonants}–\ref{tab:gemination}. As shown in Table~\ref{tab:vowels-consonants}, consonants dominate the dataset, with 32,688 occurrences compared to 16,101 vowels. This consonant-heavy profile is characteristic of Tashlhiyt phonotactics, which allows complex consonant clusters and even entirely vowelless syllables \cite{Ridouane2014}. The resulting skew towards consonantal segments highlights one of the main typological challenges for speech and text processing: syllabification, vowel insertion, and phoneme alignment are considerably less straightforward than in more vowel-rich languages.

Table~\ref{tab:pharyngeals} details the distribution of pharyngeal and uvular consonants, which are salient features of the Tashlhiyt phonemic inventory. The voiced uvular fricative /ɣ/ is the most frequent member of this class (1,193 tokens), reflecting its high lexical productivity and presence in both native and borrowed forms. The voiced pharyngeal /\textipa{E}/ and the voiceless pharyngeal /ḥ/ also occur frequently, illustrating the robust presence of pharyngeal contrasts. These sounds are typologically significant and contribute to spectral and articulatory variability that can complicate both ASR and text normalization tasks \cite{Ridouane2014, Boulaknadel2018}.

\begin{table}[t]
\centering
\begin{tabular}{l c@{\hskip 0.4cm}l c}
\hline
\textbf{Consonants} & \textbf{Count} & \textbf{Consonants} & \textbf{Count} \\
\hline
tt & 810 & ll & 465 \\
ss & 372 & nn & 397 \\
dd & 295 & mm & 150 \\
qq & 187 & kk & 182 \\
yy & 137 & gg & 139 \\
ṣṣ & 121 & ṭṭ & 111 \\
zz & 110 & cc & 101 \\
ḍḍ & 50 & bb & 48 \\
jj & 48 & ww & 46 \\
rr & 58 & \textsubdot{z}\textsubdot{z} & 41 \\
ff & 20 & pp & 13 \\
xx & 11 & ḥḥ & 5 \\
hh & 3 & ɣɣ & 1 \\
\hline
\end{tabular}
\caption{Number of occurrences of geminated consonant patterns in the standardized Tashlhiyt corpus, ranked from most to least frequent.}
\label{tab:gemination}
\end{table}

Labialization patterns, shown in Table~\ref{tab:labialization}, reveal that labialized consonants—especially \textit{g\textsuperscript{w}} and \textit{k\textsuperscript{w}}—are relatively rare but systematically attested. Their occurrence demonstrates that the corpus captures fine-grained phonological detail present in natural Tashlhiyt, where secondary articulations such as labialization and palatalization play a role in lexical contrasts. The inclusion of these forms is relevant for developing orthographic normalization schemes capable of handling diacritically marked consonants and superscript notations (e.g., \textsuperscript{w}).

Finally, Table~\ref{tab:gemination} summarizes the distribution of geminated consonant patterns. Gemination is pervasive in Tashlhiyt morphology and phonology, often signaling morphological boundaries, aspectual distinctions, or lexical contrasts. The most frequent geminates—\textit{tt}, \textit{ll}, and \textit{nn}—reflect common morphological roots and derivational patterns. Interestingly, emphatic and pharyngeal geminates (\textit{ṭṭ}, \textit{ṣṣ}, \textit{ḍḍ}, \textsubdot{z}\textsubdot{z}) are also well represented, underlining the phonological complexity that any normalization or grapheme-to-phoneme model must account for. The rarity of geminated /ɣ/ and /ḥ/ is consistent with phonotactic restrictions reported in descriptive grammars, where such patterns are typically disallowed or highly marked \cite{Ridouane2014}. Overall, the phoneme-level statistics confirm that the corpus preserves authentic phonological structures of Tashlhiyt, making it suitable for computational modeling that respects the language’s typological and articulatory specificities.

\begin{table}[t]
\centering
\begin{tabular}{l l l}
\hline
\textbf{Neo. shi} & \textbf{Ver. shi} & \textbf{English} \\
\hline
tihirit & ṭṭumubil & car \\
tasnbḍayt & lmḥkama & court \\
tusnakt & lmaṭ & mathematics \\
\hline
\end{tabular}
\caption{Examples comparing Neologism (Neo.) and Vernacular (Ver.) Tashlhiyt words.}
\label{tab:neo-ver-shilha}
\end{table}

\section{Methodology}
\label{sec:methodology}

\subsection{Orthography}

Normalizing user-generated Tashlhiyt text is challenging due to the absence of a single, universally applied writing standard. While Tifinagh is the official script promoted by IRCAM, literacy remains limited \cite{AitLaaguidKhaloufi2023}. Consequently, speakers often use non-standard Latin orthographies influenced by French and Arabic chat conventions, creating a gap between spontaneous writing and standardized forms required for NLP.

We adopted the Berber Latin script \cite{NaitZerad2011}, following IRCAM’s conventions \cite{Boukhris2008}, as it balances inclusivity and practical usability. To illustrate, consider:\\

\noindent 
\textbf{Non-standard Tashlhiyt:} \texttt{Irgha l7al ossanad, our ssn7 manikhd kolo sigh ika l7mayad.}\\

\noindent 
\textbf{Standardized Tashlhiyt:} \texttt{Irɣa lḥal ussan ad, ur ssinɣ mani ɣ d kullu siɣ ikka lḥma ad.}\\

\noindent 
\textbf{English:} \texttt{It’s been hot these days, I don’t know where all this heat is coming from.}\\
% \end{itemize}
% \end{center}

Normalization operates on several levels. At the character level, numerals and digraphs representing absent phonemes (e.g., \texttt{7}$\rightarrow$\texttt{ḥ}, \texttt{gh}$\rightarrow$\texttt{ɣ}) are standardized \cite{Kessai2018}. Vowel representations influenced by French orthography (\texttt{o}, \texttt{ou}$\rightarrow$\texttt{u}) and inconsistent word boundaries (e.g., \texttt{ossanad}$\rightarrow$\texttt{ussan ad}) are corrected systematically.  

At the morpho-syntactic level, dialectal and grammatical variation requires linguistic normalization. For example, \texttt{our ssn7}$\rightarrow$\texttt{ur ssinɣ} involves (a) replacing the regional suffix \texttt{-ḥ} (\texttt{-7}) with \texttt{-ɣ} \cite{Chaker2019}, and (b) enforcing the negative form of the verb \texttt{ssinɣ} under \texttt{ur} \cite{Bensoukas2009}.  

Finally, phonological processes such as assimilation and fusion are resolved to preserve canonical forms. For instance, \texttt{ayt dari}$\rightarrow$\texttt{ayddari} (voicing assimilation) and \texttt{tkkst tt}$\rightarrow$\texttt{tkksst} illustrate phonetic spellings requiring morphophonemic normalization \cite{Boukhris2008}.  

\subsection{Vocabulary}

A major challenge lies in reconciling standardized Amazigh lexicons—often prescriptive and neologism-rich—with contemporary spoken Tashlhiyt \cite{aitouguengay2010standardised, bouhjar2008amazigh}. Many official corpora employ terms unfamiliar to most speakers \cite{idhssaine2023critical}. Our approach is descriptive, prioritizing lexical authenticity and modern usage.  

We excluded artificial neologisms lacking community adoption, retaining common loanwords from Darija, French, and English, which form part of the living lexicon \cite{Soulaimani2016}. Table~\ref{tab:neo-ver-shilha} illustrates typical contrasts between neologisms and their vernacular equivalents.

Archaic forms were included only when still widely understood, ensuring that the resulting corpus reflects real communicative usage rather than an idealized linguistic norm. This approach aims to create a dataset both linguistically sound and practically relevant for NLP applications.

\section{Experimental Setup}
\label{sec:experimental}

\subsection{Dataset}

\begin{table}[t]
\centering
\begin{tabular}{l l l}
\hline
\textbf{Arc. shi} & \textbf{Con. shi} & \textbf{English} \\
\hline
tawuri & lxdmt & work / function \\
taḍḍanga & lmuja & wave \\
arqqas & ṛṛasul & messenger \\
\hline
\end{tabular}
\caption{Examples of archaic (Arc.) Tashlhiyt terms and their contemporary (Con.) equivalents used in the corpus.}
\label{tab:archaic-modern}
\end{table}

For the evaluation of large language models on Tashlhiyt orthography normalization, we rely on the manually standardized subset of 1,500 parallel sentence pairs from the DATASHI corpus described in Section~\ref{sec:corpus}. In addition, a carefully curated set of 30 representative sentences—distinct from the 1,500-sentence evaluation subset—was selected from the broader corpus for few-shot prompting (see Subsection~\ref{subsec:prompting}). These sentences were manually standardized and selected to maximize orthographic diversity, covering a broad spectrum of graphemic correspondences, character-level substitutions, and morphophonemic alternations characteristic of non-standard Tashlhiyt writing.

\subsection{Large Language Models}

\begin{table*}[t]
\centering
\begin{tabular}{lcccc}
\hline
\multirow{2}{*}{\textbf{Model}} & \multicolumn{2}{c}{\textbf{Zero-Shot}} & \multicolumn{2}{c}{\textbf{Few-Shot}} \\
\cline{2-5}
 & \textbf{WER (\%) $\downarrow$} & \textbf{LD $\downarrow$} & \textbf{WER (\%) $\downarrow$} & \textbf{LD $\downarrow$} \\
\hline
Claude-Sonnet-4.5    & 47.5 & 4.86 & 43.1 & 4.43 \\
Gemini-2.5-Pro  & \textbf{37.8} & \textbf{3.94} & \textbf{35.5} & \textbf{3.65} \\
GPT-5       & 51.9 & 5.60 & 48.7 & 5.25 \\
Mistral-Large-2411     & 58.2   & 6.16   & 54.7 & 5.80 \\
Qwen3-Max   & 63.4 & 7.46 & 56.8 & 5.99 \\
\hline
\end{tabular}
\caption{Average normalization performance across models in zero-shot and few-shot settings, measured using WER and LD.}
\label{tab:llm-eval}
\end{table*}

\begin{table*}[t]
\centering
\setlength{\tabcolsep}{4pt}
\renewcommand{\arraystretch}{1.1}
\begin{tabular}{lccccc}
\hline
\textbf{Category} & \textbf{Claude-Sonnet-4.5} & \textbf{Gemini-2.5-Pro} & \textbf{GPT-5} & \textbf{Mistral-Large-2411} & \textbf{Qwen3-Max} \\
\hline
Emphatics     & 271  & 188  & 307  & 360  & 317 \\
Pharyngeals   & 10   & 10   & 11   & 17   & 15  \\
Uvulars       & 8    & 11   & 9    & 12   & 13  \\
Labialization & 125  & 95   & 132  & 138  & 128 \\
Gemination    & 2962 & 2144 & 3304 & 4056 & 3738 \\
\hline
\end{tabular}
\caption{Total \textbf{deletions} by phonological category across models. Lower counts indicate fewer deletion errors relative to the reference.}
\label{tab:phonological-deletions}
\end{table*}

To establish a strong baseline for Tashlhiyt, we evaluated the performance of several state-of-the-art Large Language Models (LLMs). The objective was to measure their intrinsic ability to handle the complex orthographic, morphological, and phonological variations present in our dataset without any model fine-tuning.

%Our selection of models\footnote{The code used for preprocessing, normalization, and evaluation will be made publicly available on GitHub at the camera-ready stage to ensure full transparency and reproducibility.} 

Our selection of models was designed to cover the leading architectures from major AI research labs, representing the current frontier of language understanding and generation capabilities. The following five models were used in our experiments: GPT-5 \cite{openai2025gpt5}, OpenAI's flagship multimodal model known for setting industry benchmarks in reasoning and multilingual tasks; Claude-Sonnet-4.5 \cite{anthropic2025claudesonnet4_5}, Anthropic's latest model recognized for its speed and strong performance in rule-intensive instruction following; Gemini-2.5-Pro \cite{comanici2025gemini2_5}, Google's state-of-the-art model distinguished by its massive context window and advanced multilingual reasoning; Mistral Large 2411 \cite{mistral2024large2411}, the top-tier proprietary model from Mistral AI, highly regarded for its powerful reasoning and multilingual fluency; and Qwen3 Max \cite{qwen3max}, a top-performing open model from Qwen that has demonstrated capabilities competitive with leading proprietary systems.

\subsection{Prompting}
\label{subsec:prompting}

Our prompting strategy was designed to rigorously constrain the LLMs' outputs and ground their behavior in established linguistic rules. The prompt itself is multi-faceted: it opens with a direct instruction to normalize the input, immediately followed by an exhaustive character set (\texttt{a, b, c, ḍ...}) to prevent invalid orthography. The core of the prompt details the specific IRCAM-based rules for handling key phonological and morphological features of Tashelhiyt, including labialization (\textsuperscript{w}), emphatic/pharyngeal consonants (\texttt{ṭ}, \texttt{ḍ}, \texttt{ḥ}, \textipa{E}), gemination, and vowel standardization. While this prompt provided the full context for the \textbf{zero-shot} evaluation, in the \textbf{few-shot} setting it framed the task, allowing the models to generalize from the provided examples based on these explicit rules.%\footnote{The full text of the prompt will be provided in Appendix at the camera-ready stage.}.

\subsection{Evaluation metrics}

Normalization performance was evaluated using Word Error Rate (WER) and Levenshtein distance (LD). WER provides a token-level measure of insertion, deletion, and substitution errors between predicted and reference sentences, while the LD offers a complementary character-level metric sensitive to finer orthographic variations typical of Tashlhiyt spelling.

During preliminary experiments, some LLM outputs introduced characters not included in the orthographic inventory specified in the prompt (e.g., \texttt{č}, \texttt{š}). These cases were counted as insertion errors but are not reported separately in the analysis.

\section{Results and Discussion}
\label{sec:results}

\subsection{Overall Normalization Performance}

Table~\ref{tab:llm-eval} presents the average normalization performance of all evaluated models under both zero-shot and few-shot settings, using two complementary metrics: the WER and the LD. 

In the zero-shot setting, Gemini-2.5-Pro clearly outperformed all other models, achieving the lowest WER (37.8\%) and LD (3.94). Claude-Sonnet-4.5 followed with moderate results (WER~47.5\%, Lev.~4.86), while GPT-5 and Mistral showed higher error rates (WER~51.9\% and 58.2\%, respectively). Qwen3-Max displayed the weakest performance, with both the highest WER (63.4\%) and LD (7.46), suggesting difficulty in handling the morphological and orthographic variability of the corpus when no prior examples were provided.

In the few-shot setting, which incorporates a small number of demonstration examples, all models improved to varying degrees. Gemini-2.5-Pro again achieved the best overall performance (WER~35.5\%, Lev.~3.65), confirming its robustness and adaptability to the normalization task. Claude-Sonnet-4.5 and GPT-5 also benefited from the few-shot configuration, reducing their WERs to 43.1\% and 48.7\%, respectively. Mistral and Qwen3-Max showed improvements, though their relative ranking remained unchanged compared to the zero-shot condition.
Overall, these results highlight Gemini-2.5-Pro as the most reliable model for Tashlhiyt normalization, both in zero-shot and few-shot contexts. The consistent improvement across models in the few-shot setup also demonstrates the benefit of in-context learning for normalization tasks involving low-resource or morphologically rich languages.

\subsection{Fine-Grained Phonological Error Analysis}

\begin{table}[t]
\centering
\begin{tabular}{lccc}
\hline
\textbf{Model} & \textbf{D} & \textbf{S} & \textbf{I} \\
\hline
Claude-Sonnet-4.5   & 3000 & 3215 & 1452 \\
Mistral-Large-2411    & 4131 & 4405 & \textbf{1293} \\
Gemini-2.5-Pro & \textbf{2182} & \textbf{2516} & 1818 \\
Qwen3-Max & 3801 & 4673 & 4672 \\
GPT-5      & 3357 & 3806 & 1732 \\
\hline
\end{tabular}
\caption{Total counts of deletions, substitutions, and insertions (D, S, I) across the Tashlhiyt corpus for each model output in few-shot setting.}
\label{tab:edit-ops-models}
\end{table}

\begin{table*}[t]
\centering
\setlength{\tabcolsep}{4pt}
\renewcommand{\arraystretch}{1.1}
\begin{tabular}{lccccc}
\hline
\textbf{Category} & \textbf{Claude-Sonnet-4.5} & \textbf{Gemini-2.5-Pro} & \textbf{GPT-5} & \textbf{Mistral-Large-2411} & \textbf{Qwen3-Max} \\
\hline
Emphatics     & 1146 & 686  & 1183 & 1165 & 1450 \\
Pharyngeals   & 41   & 34   & 250  & 576  & 391  \\
Uvulars       & 103  & 37   & 283  & 570  & 796  \\
Labialization & 59   & 26   & 26   & 14   & 22   \\
Gemination    & 3084 & 2380 & 3626 & 4200 & 4470 \\
\hline
\end{tabular}
\caption{Total \textbf{substitutions} by phonological category across models. Lower counts indicate fewer substitution errors relative to the reference.}
\label{tab:phonological-substitutions}
\end{table*}

\begin{table*}[t]
\centering
\setlength{\tabcolsep}{4pt}
\renewcommand{\arraystretch}{1.1}
\begin{tabular}{lccccc}
\hline
\textbf{Category} & \textbf{Claude-Sonnet-4.5} & \textbf{Gemini-2.5-Pro} & \textbf{GPT-5} & \textbf{Mistral-Large-2411} & \textbf{Qwen3-Max} \\
\hline
Emphatics     & 10   & 14   & 12   & 0    & 11  \\
Pharyngeals   & 9    & 10   & 30   & 15   & 109 \\
Uvulars       & 3    & 3    & 16   & 1    & 46  \\
Labialization & 14   & 2    & 15   & 1    & 0   \\
Gemination    & 1366 & 1714 & 1626 & 1185 & 4296 \\
\hline
\end{tabular}
\caption{Total \textbf{insertions} by phonological category across models. Lower counts indicate fewer insertion errors relative to the reference.}
\label{tab:phonological-insertions}
\end{table*}

In this section, we conduct a detailed analysis of edit operations, decomposed into deletions, substitutions, and insertions, as summarized in Table~\ref{tab:edit-ops-models}. To refine this analysis, Tables~\ref{tab:phonological-deletions}--\ref{tab:phonological-insertions} present a phonological decomposition of these edit operations, showing how each model performs with respect to specific phonological categories, namely emphatics, pharyngeals, uvulars, labialized consonants, and geminates.

\subsubsection*{Deletions} Overall, the results indicate that Gemini-2.5-Pro consistently produces the lowest deletion counts across nearly all phonological classes, confirming the robustness already observed at the global level. Claude-Sonnet-4.5 and GPT-5 follow closely, showing moderate performance particularly on emphatic and uvular consonants. In contrast, Mistral and Qwen3-Max display substantially higher deletion frequencies, particularly in geminated segments, where deletions are an order of magnitude greater than in other categories.

The concentration of deletion errors in gemination and emphatic consonants reflects the models’ sensitivity to non-concatenative morphological and morphophonological structures characteristic of Amazigh languages. These findings emphasize the importance of subword-level and orthography-aware modeling in improving normalization accuracy, particularly for morphologically rich and under-resourced languages such as Tashlhiyt.

\subsubsection*{Substitutions} Substitution errors are generally more frequent than deletions or insertions across all models, indicating that normalization discrepancies tend to involve incorrect replacements rather than omissions or additions. Overall, Gemini-2.5-Pro exhibits the lowest substitution occurrences across all phonological classes, followed by Claude-Sonnet-4.5 and GPT-5, while Mistral and Qwen3-Max display substantially higher rates.

A closer examination by phonological category (Table~\ref{tab:phonological-substitutions}) reveals that gemination dominates substitution errors for all models, with Mistral and Qwen3-Max producing the largest counts (4,200 and 4,470, respectively). In contrast, Gemini-2.5-Pro maintains the lowest gemination substitutions (2,380), showing better preservation of consonant length contrasts. For emphatic consonants, substitution errors remain relatively stable across models, with the lowest values observed for Gemini-2.5-Pro (686) and the highest for Qwen3-Max (1,450). In pharyngeal and uvular categories, Claude-Sonnet-4.5 and Gemini-2.5-Pro show minimal confusion, whereas Mistral and GPT-5 exhibit more frequent replacements, suggesting weaker robustness in handling marked phonological features.

These results indicate that substitution errors are not uniformly distributed but rather concentrated in phonologically marked segments such as emphatics, pharyngeals, and geminates. While Gemini-2.5-Pro demonstrates the most balanced behavior across categories, Mistral and Qwen3-Max tend to over-normalize or distort these phonologically complex orthographic forms, highlighting their limitations in capturing segmental distinctions reflected in Tashlhiyt morphology and orthography.

\subsubsection*{Insertions} Insertion errors are generally less frequent than deletions and substitutions, indicating that normalization discrepancies arise predominantly from omissions or replacements rather than from added segments.

Across models, Mistral exhibits the lowest overall insertion counts, followed closely by Claude-Sonnet-4.5 and Gemini-2.5-Pro, both of which show relatively controlled insertion behavior across categories. In contrast, Qwen3-Max displays markedly higher insertion rates, particularly for gemination (4,296) and pharyngeal segments (109). GPT-5 produces moderate insertion levels, with errors concentrated in geminated and labialized consonants, suggesting some instability in handling orthographic repetition and segmental marking.

The predominance of insertions in geminated consonants again underscores the models’ limited capacity to preserve consonant length contrasts—a salient morphological and orthographic cue in Tashlhiyt. Occasional over-generation of pharyngeal and uvular consonants in lower-performing models may reflect inconsistent mappings of marked phonological segments to their normalized orthographic equivalents. Overall, insertion patterns corroborate the trends observed for deletions and substitutions: Mistral, Gemini 2.5, and Claude 4 maintain relatively stable normalization behavior, while Qwen3-Max exhibits the greatest redundancy and inconsistency across phonological categories.

\section{Conclusion}
\label{sec:conclusion}

In this paper, we presented \textbf{DATASHI}, the first parallel English–Tashlhiyt corpus designed for orthography normalization and broader Amazigh NLP development. By combining user-generated and expert-standardized texts, it captures real orthographic variation while providing a consistent benchmark for normalization and translation tasks. 

Evaluation with state-of-the-art LLMs showed that Gemini-2.5-Pro achieves the best normalization accuracy and most stable phonological coverage, confirming that foundation models can generalize to low-resource settings but still face challenges with gemination and emphatic contrasts. 

Beyond its textual contribution, DATASHI establishes a scalable foundation for multimodal extensions—particularly speech alignment and pronunciation modeling. Future work will expand the corpus with read-speech data, fine-tune multilingual models on normalized text, and integrate DATASHI into cross-Amazigh benchmarks. Together, these steps aim to advance resource equity and reproducible evaluation for Amazigh languages.

\section{Ethical Statement}
\label{sec:ethics}

All participants gave informed consent prior to participation. Participation was voluntary, and contributors were informed about the research objectives before data collection. No personal data were collected. The corpus respects speaker privacy, cultural representation, and linguistic diversity. Data and scripts will be openly released for reproducible research, following the LREC Ethical Charter and ensuring responsible, community-centered development of Tashlhiyt language resources.

During preliminary experiments, we observed that several LLMs occasionally introduced characters not included in the orthographic inventory specified in the prompt, such as \texttt{č} or \texttt{š}. Because such outputs violate the explicitly defined character set and reflect orthographic conventions outside the scope of the normalization task, they were excluded from the error analysis reported in this paper.

\section{Bibliographical References}\label{sec:reference}

\bibliographystyle{lrec2026-natbib}
\bibliography{lrec2026}

% \section{Language Resource References}
\label{lr:ref}
\bibliographystylelanguageresource{lrec2026-natbib}
\bibliographylanguageresource{languageresource}

\end{document}